\documentclass{article}

\usepackage[preprint]{neurips_2025}
\usepackage{graphicx}
\usepackage{amsmath}
\usepackage{algorithm, algpseudocode}
\usepackage{multirow}
\usepackage{array}
\usepackage{pifont}

\usepackage[utf8]{inputenc} 
\usepackage[T1]{fontenc}    
\usepackage{hyperref}       
\usepackage{url}            
\usepackage{booktabs}       
\usepackage{amsfonts}       
\usepackage{nicefrac}       
\usepackage{microtype}      
\usepackage{xcolor}         

\title{Focus Your Attention: Towards Data-Intuitive Lightweight Vision Transformers}

\author{%
  Suyash Gaurav \thanks{These authors contributed equally and share first authorship.} \\
  Tokyo International University\\
  Tokyo, Japan\\
  \And 
  Muhammad Farhan Humayun \footnotemark[1]\\
  University of Turku \\
  Turku, Finland\\
  \And
  Jukka Heikkonen \\
  University of Turku \\
  Turku, Finland\\
  \And 
  Jatin Chaudhary \\
  University of Turku \\
  Turku, Finland\\
  \texttt{jatin.chaudhary@utu.fi} \\
}

\begin{document}

\maketitle

\begin{abstract}
  The evolution of Vision Transformers has led to their widespread adaptation to different domains. Despite large-scale success, there remain significant challenges including their reliance on extensive computational and memory resources for pre-training on huge datasets as well as difficulties in task-specific transfer learning. These limitations coupled with energy inefficiencies mainly arise due to the computation-intensive self-attention mechanism. To address these issues, we propose a novel \textbf{Super-Pixel Based Patch Pooling (SPPP)} technique that generates context-aware, semantically rich, patch embeddings to effectively reduce the architectural complexity and improve efficiency. Additionally, we introduce the \textbf{Light Latent Attention (LLA)} module in our pipeline by integrating latent tokens into the attention mechanism allowing cross-attention operations to significantly reduce the time and space complexity of the attention module. By leveraging the data-intuitive patch embeddings coupled with dynamic positional encodings, our approach adaptively modulates the cross-attention process to focus on informative regions while maintaining the global semantic structure. This targeted attention improves training efficiency and accelerates convergence. Notably, the SPPP module is lightweight and can be easily integrated into existing transformer architectures. Extensive experiments demonstrate that our proposed architecture provides significant improvements in terms of computational efficiency while achieving comparable results with the state-of-the-art approaches, highlighting its potential for energy-efficient transformers suitable for edge deployment. (The code is available on our GitHub repository: https://github.com/zser092/Focused-Attention-ViT).
\end{abstract}

\section{Introduction}
The self-attention mechanism which forms the core of the transformer architecture was originally designed to handle a variety of Natural Language Processing (NLP) tasks such as machine translation~\cite{NIPS2017_3f5ee243}, sentiment classification~\cite{DBLP:journals/corr/LinFSYXZB17,ambartsoumian-popowich-2018-self} sentence prediction~\cite{DBLP:journals/corr/abs-1810-04805} reading comprehension~\cite{NEURIPS2019_dc6a7e65} etc. The mechanism has been further adapted to advanced reasoning models such as Generative Pretrained Transformer (GPT)~\cite{Radford2018ImprovingLU} and DeepSeek~\cite{deepseekai2025deepseekv3technicalreport}. The principal advantage of the self-attention mechanism is its ability to effectively capture the long-range dependencies more efficiently as compared to its old rival i.e. the recurrent neural networks (RNNs). Concretely, self-attention enables the model to weigh the relevance of different tokens in a sequence relative to each other, regardless of their position. 

\begin{figure}[hbt!]
    \centering
    \includegraphics[width=12.5cm]{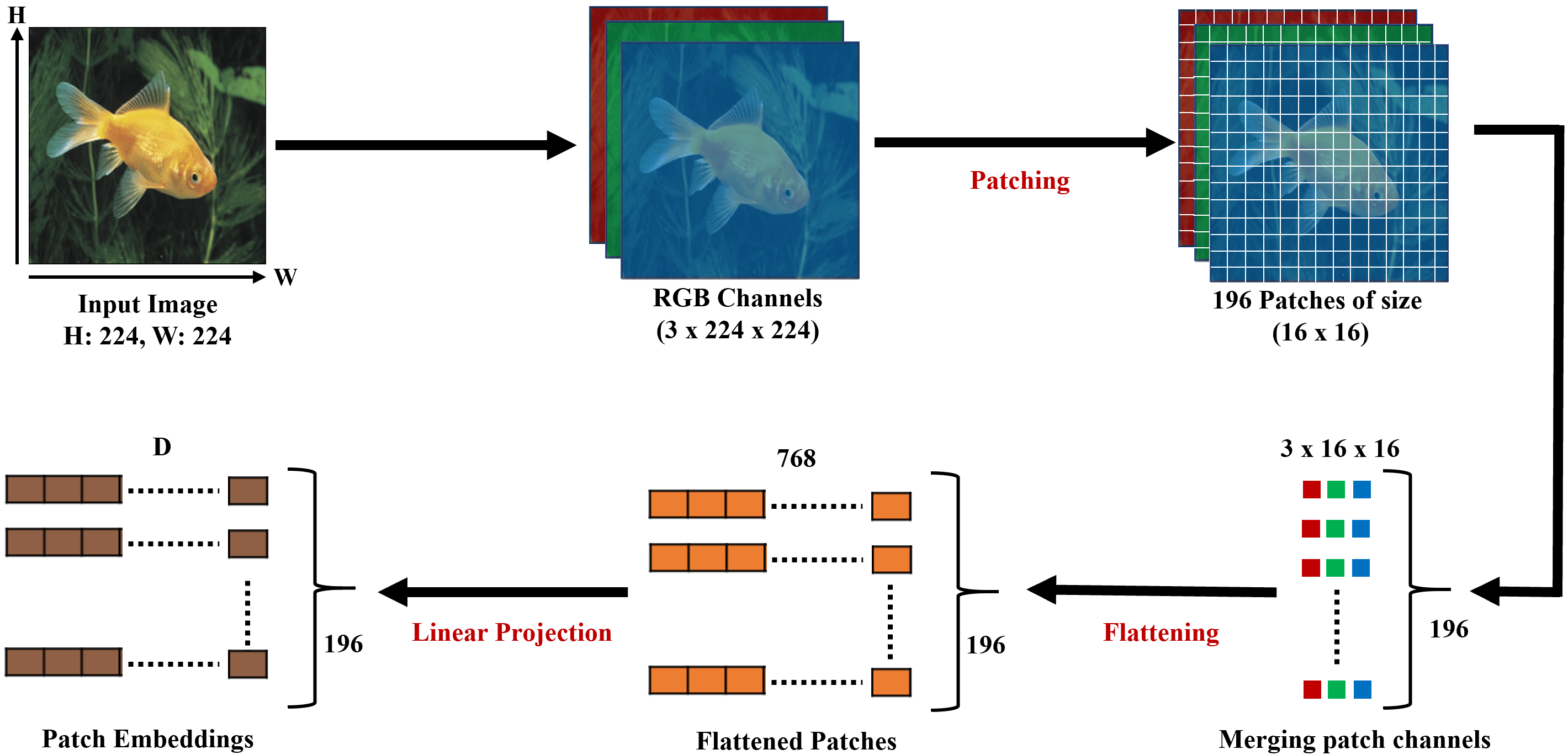}
    \caption{Illustration of the Patch Embedding Module for Vision Transformers}
    \label{fig1}
\end{figure}

The adaptation of this approach towards Computer Vision tasks, first proposed by Dosovitskiy et.al, ~\cite{DBLP:conf/iclr/DosovitskiyB0WZ21} allows the processing of 2D input image patches in a similar fashion compared to the word tokens in NLP tasks. Figure\ref{fig1}illustrates this process in Vision Transformers (ViT) from the input image to the patch embeddings which are directly passed on to the multi-head self-attention modules. To exemplify, we use a resampled goldfish image from the ImageNet dataset to show the intermediate outputs generated during the embedding process. Considering the input image is: \( \mathbf{X} \in \mathbb{R}^{H \times W \times C} \).

In the provided example, the input image is resampled to 224 x 224 pixels and contains 3 color channels. It is then divided into 196 equal sized, non-overlapping patches of 16 x 16 using the following equation:
\begin{equation}
N = \frac{H \times W}{P^2}
\end{equation}
\noindent

Here, $N$ denotes the number of patches (excluding the class token), $H$ and $W$ correspond to the height and width of the input image, respectively, and $P$ is the side length of each square patch (i.e., assuming square patches of size $P\times P$).

Each patch is flattened to form a vector, 
$\mathbf{x}_i \in \mathbb{R}^{P^2 \cdot C}, \quad \text{for } i = 1, \dots, N$. In the provided example, we have 196 different vectors of size $16\times 16\times 3 = 768$. Each flattened patch vector \( \mathbf{x}_i \) is projected to a lower-dimensional embedding space using a learnable linear transformation: \( \mathbf{E} \in \mathbb{R}^{(P^2 \cdot C) \times D} \), where \( D \) is the transformer embedding dimension:
\begin{equation}
\mathbf{z}_i = \mathbf{x}_i \mathbf{E} \in \mathbb{R}^{D}
\end{equation}
A key observation here is that the number of final patch embeddings scales linearly with the number of input image patches which in turn depends on the input image resolution and the individual patch size. In the subsequent multi-headed self-attention, each patch embedding is correlated against one another resulting in quadratic complexity i.e., $O(N^2)$. Considering the example as illustrated in Figure 1, there are $196$ patch embeddings which amount to $(196)^2\approx38k$ pairwise correlation computations within the self-attention module.

We introduce robust techniques to effectively reduce the embedding dimensions by intelligently fusing the input patches based on the inherent semantic structure. This in turn reduces the computational overhead of the self-attention module, while preserving and propagating rich semantic information to the deeper layers of the network to facilitate convergence and more efficient training. The main contributions of our work are as follows:   

\begin{itemize}
\item We propose the novel Super-Pixel Based Patch Pooling (SPPP) algorithm to intelligently reduce the number of patch embeddings which are fed into the attention mechanism by merging the input image patches based on their intrinsic semantic properties.  
\item We propose replacing the multi-headed attention modules with the \textbf{novel} Light Latent Attention (LLA) mechanism in the ViT architecture, enabling efficient cross-attention operations with latent tokens.
\item Our architectural modifications significantly reduce the time and space complexity of the attention mechanism which is the major bottleneck in the ViT architecture in terms of computational and temporal requirements.
\item We conduct extensive experiments and ablation studies on various public benchmark datasets to validate the effectiveness of our proposed techniques. 
\end{itemize}

\section{Background and Related Work}
Vision Transformers (ViTs) have shown remarkable success across various tasks from image classification \cite{DBLP:conf/iclr/DosovitskiyB0WZ21, pmlr-v139-touvron21a, 9710580, technologies13010032} to more complex tasks including segmentation, \cite{9578646, NEURIPS2021_64f1f27b, NEURIPS2021_3bbfdde8, THISANKE2023106669} object detection, \cite{10.1007/978-3-030-58452-8_13, liu2022dabdetr, 10657220, Zhang2025} and video understanding \cite{9710415, pmlr-v139-bertasius21a, article123}. Similarly, a large number of domain-specific studies utilizing transformer architectures also exist, such as medical image analysis \cite{9706678, HE202359, Aburass2025}, remote sensing \cite{9627165, 10147258}, pose estimation \cite{9674785} and action recognition \cite{10030830} etc. Since the groundbreaking approach of replacing the traditional convolutional operations entirely with attention-based architectures proposed by Dosovitskiy et.al~\cite{DBLP:conf/iclr/DosovitskiyB0WZ21}, the ViT architecture has been studied vastly proposing improved, more robust variants of the originally proposed model. For instance, to mitigate the heavy computational burden, several works have proposed architectural modifications aimed at improving efficiency. DeiT \cite{pmlr-v139-touvron21a} introduced knowledge distillation to train ViTs on smaller datasets without sacrificing accuracy. T2T-ViT \cite{9710747} employed a tokens-to-token transformation that progressively aggregates information before applying self-attention. Similarly, MobileViT \cite{mehta2022mobilevit} integrated convolutions into the transformer pipeline to enhance locality and reduce computational cost, making ViTs more viable for mobile settings. Other works like PoolFormer \cite{9879612} and PiT \cite{9710548} introduced hierarchical token reduction strategies and spatial pooling mechanisms to reduce input sequence length progressively. LeViT \cite{9711161} leveraged downsampling and hybrid convolutions to strike a better latency-accuracy trade-off. These variants highlight a trend toward building lightweight and scalable ViTs, but they often compromise the richness of patch representations or spatial context.

The self-attention mechanism scales quadratically with input sequence length. To address this, Linformer \cite{DBLP:journals/corr/abs-2006-04768} approximated self-attention with low-rank projections, while Nyströmformer \cite{Xiong_Zeng_Chakraborty_Tan_Fung_Li_Singh_2021} introduced kernel-based and landmark-based approximations. Swin Transformer\cite{9710580} introduced a hierarchical architecture with shifted window attention to improve scalability and efficiency in vision tasks. Its successor, Swin Transformer V2 \cite{9879380}, further enhanced training stability and generalization through log-scaled relative position bias and scaled cosine attention, enabling deployment at a billion-parameter scale. Light Vision Transformer \cite{9879515} proposed separate enhanced self-attention mechanisms for low and high-level features to improve efficiency. DynamicViT \cite{NEURIPS2021_747d3443} proposed learning to prune tokens during inference, enabling adaptive computation without retraining. CrossViT \cite{9711309} demonstrated the value of multi-scale token fusion using cross-attention, whereas other works like TokenLearner \cite{NEURIPS2021_6a30e32e} and AdaViT \cite{9879366} introduced dynamic token selection and routing to focus computation on informative regions, similar in spirit to the dynamic attention strategy we explore. ShiftAddViT \cite{NEURIPS202369c49f75} reduces the computational cost of Vision Transformers by reparameterizing attention and MLPs using shift and addition operations, enabling significant GPU latency and energy savings without retraining from scratch. More recently, Look Here Vision Transformer \cite{fuller2024lookhere} proposed a novel position encoding method via 2D attention masks to improve translation equivariance and attention head diversity. Similarly, MicroViT proposed \cite{setyawan2025microvitvisiontransformerlow} an efficient single-head attention mechanism to reduce the computational complexity of the overall ViT architecture.

Superpixels have been leveraged for context-aware representations in traditional vision tasks, but their integration into transformers remains underexplored. Superpixel-based attention \cite{10341519} and patch grouping strategies have shown promise in enhancing spatial coherence while reducing token count. Our proposed SPPP module builds on this insight by using semantically aligned patch pooling to guide attention and reduce computational overhead.

\section{Methodology}

\begin{figure}[hbt!]
    \centering
    \includegraphics[width=7cm]{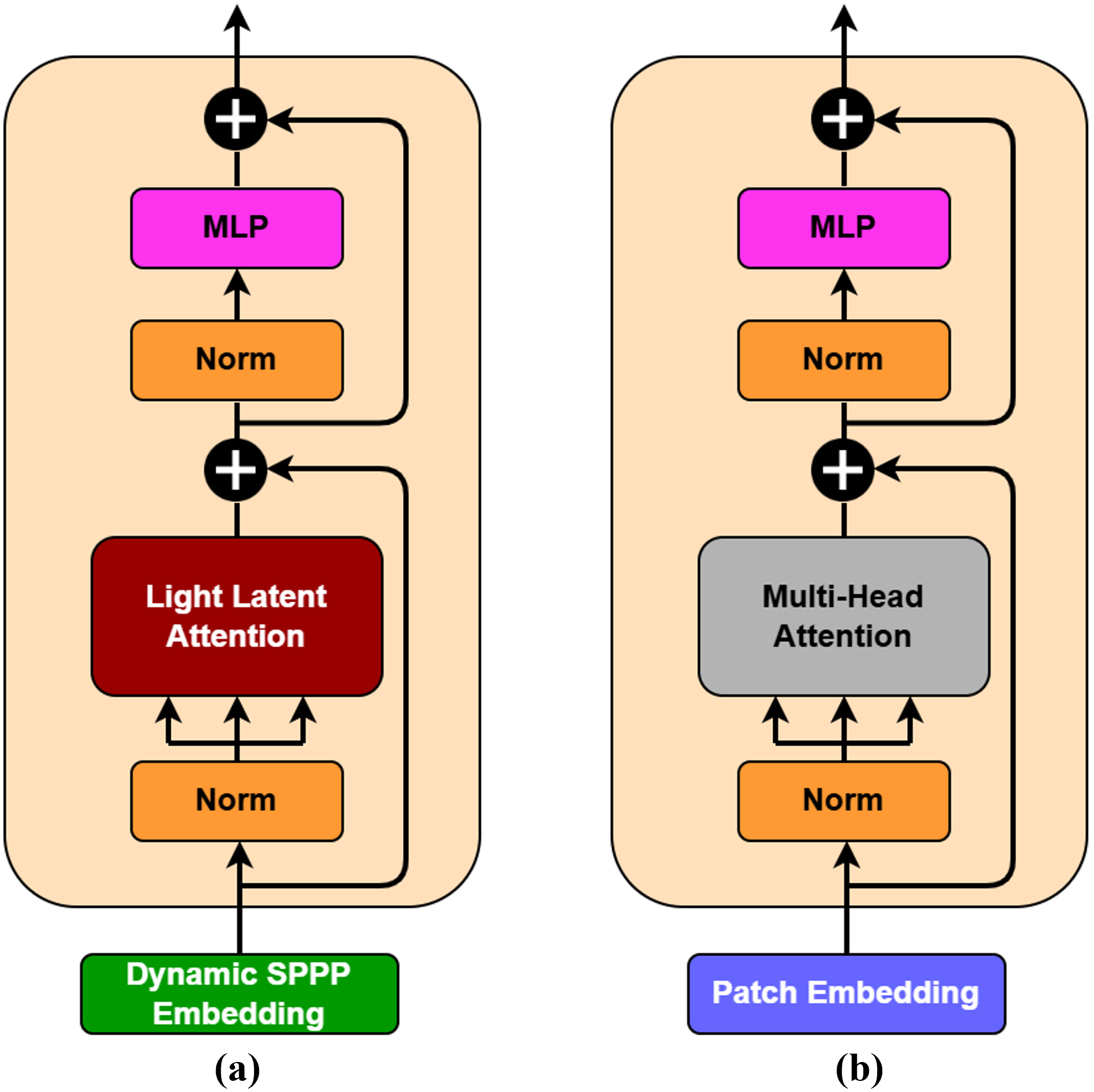}
    \caption{Key architectural differences: (a) Proposed ViT with Focused Attention, (b) Original ViT.}
    \label{fig2}
\end{figure}

In this section, we discuss the proposed architectural changes to the ViT in detail. Figure\ref{fig2} demonstrates the key differences of the proposed architecture compared with the original ViT. The first major difference is that the regular, fixed grid patch embedding module is replaced by our newly proposed, dynamic SPPP module which generates compact, high-semantic embeddings. The second key difference is the modification in the attention module. The regular multi-head attention module is replaced by the novel LLA which uses latent tokens to reduce the computational overhead of the attention module, making the overall architecture light-weight. The details of the newly proposed modules are discussed in the following sub-sections. 

\subsection{Super-Pixel Based Patch Pooling (SPPP)}
\begin{figure}[hbt!]
    \centering
    \includegraphics[width=12.5cm]{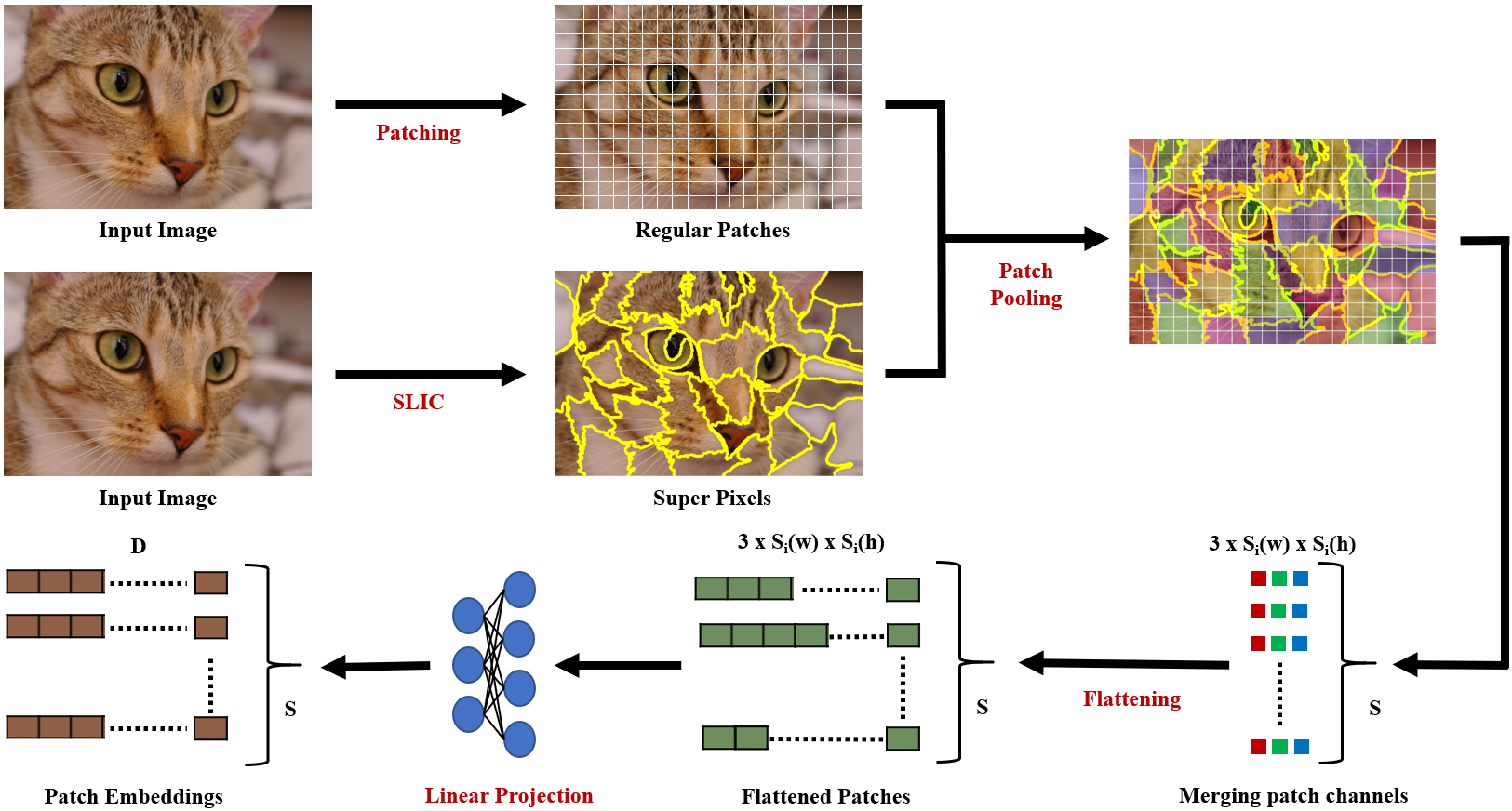}
    \caption{Workflow of the SPPP module.}
    \label{fig3}
\end{figure}

The main idea proposed in the SPPP module is to generate dynamic and semantically rich patch embeddings instead of the regular static embeddings used in the ViT. Figure\ref{fig3} provides an overview of the various steps involved in the SPPP module.

Let us break down the SPPP module. First, we divide the input image into regular equal-sized patches according to the criteria mentioned in Eq. 1. In parallel, we apply the Simple Linear Iterative Clustering (SLIC) algorithm \cite{6205760}, to group the input image into regions with coherent semantic properties called Super-pixels. This can be conceptualized as puzzle pieces that naturally follow the semantic features such as color, texture and edges etc. and typically comprise a meaningful part of the scene. The SLIC performs searching in  a fixed search space grid of $2S \times 2S$ to improve computational efficiency, such that:
\begin{equation}
    S = \sqrt{\left(\frac{N}{K}\right)} 
\end{equation}

Here, $S$ is the sampling interval, $N$ is the number of pixels in the image, and $K$ is the number of clusters. The strength of the SLIC algorithm is that it performs clustering in 5D space combining the distance measures in color and spatial dimensions into a single metric using the following equation:
\begin{equation}
    D = \sqrt{({d_c})^2 + \left(\frac{d_s}{S}\right)^2 m^2} 
\end{equation}

Here, $d_c$ is the distance in color space, $d_s$ is the spatial distance, and $m$ is the compactness factor which balances color similarity and spatial proximity. A larger value of $m$ means that more importance is given to spatial proximity, whereas a lower value of $m$ signifies more value for color and tone similarity. Once we have the initial super-pixels from the SLIC algorithm, we overlay the regular patch grids on the super-pixels and compute the overlap of all the patches with corresponding super-pixels. The following equation sums up this operation:

\begin{equation}
    O_{i,j}
= \frac{\bigl|\,S_i \cap P_j\bigr|}{|P_j|}
= \frac{\displaystyle\sum_{(x,y)\in P_j}
\mathbf{1}\bigl((x,y)\in S_i\bigr)}
{16\times16}
\end{equation}
Here, $S_i$ is the $i^{th}$ super-pixel produced by the SLIC algorithm, $P_j$ is the $j^{th}$ regular patch from the grid (e.g., 16×16 patch size), $S_i \cap P_j$ is the overlap region between the $i^{th}$ super-pixel and the $j^{th}$ regular patch, $|P_j|$ is the area (number of pixels) of the regular patch, here assumed to be $16 \times 16$. The indicator function $\mathbf{1}((x,y) \in S_i)$ equals 1 if the pixel at $(x,y)$ belongs to super-pixel $S_i$ and 0 otherwise. We merge patches j,k if there exists an i such that both \(O_{i,j}\) and \(O_{i,k}\) exceed a threshold \(\tau\). 

Finally, we proceed by flattening the RGB color channels corresponding to each super patch to form flattened vectors. It is worth noting that in our case since the size of each super patch varies, our flattened vectors also have varying lengths depending on the number of pixels included in the corresponding super patches. The flattened patch vectors are then linearly projected to fixed-sized super-patch embeddings as demonstrated in Figure\ref{fig3}. This gives us one embedding per super patch, rather than many small, noisy patch embeddings. It is also notable that the number of final super patch embeddings $S$, each representing a meaningful region of the image is considerably less as compared to the initial patch embeddings $N$ as shown in Figure\ref{fig1}, which in turn leads to less FLOPS and hence, improved spatial and temporal computational efficiency.

To preserve the spatial awareness with respect to the super patch embeddings, we compute new dynamic positional encodings that are consistent with the reduced set of tokens based on the centroid of each super-pixel such that: 
 \[
  c_x^r = \frac{1}{|S_r|} \sum_{(x,y) \in S_r} x, \quad c_y^r = \frac{1}{|S_r|} \sum_{(x,y) \in S_r} y,
  \]
  then:
  \[
  PE_r = \text{MLP}\left( \frac{c_x^r}{W}, \frac{c_y^r}{H} \right) \in \mathbb{R}^D,
  \]

Here, $S_r$ denotes the set of pixels belonging to the $r^{th}$ super-pixel and $|S_r|$ is the number of pixels in $S_r$, i.e., the cardinality of the set. The variables $(x, y)$ represent pixel coordinates within $S_r$, while $c_x^r$ and $c_y^r$ are the x- and y-coordinates of the centroid of superpixel $S_r$, respectively. The terms $W$ and $H$ refer to the width and height of the input image in pixels. The expressions ${c_x^r}/{W}$  and  ${c_y^r}/{H}$  correspond to the normalized centroid coordinates in the range $[0, 1]$. The function $\text{MLP}$ denotes a Multi-Layer Perceptron, which maps these normalized coordinates to a positional encoding vector $PE_r \in \mathbb{R}^D$, where $D$ is the embedding dimension used in the transformer model.

Following the flattening stage, to maintain spatial coherence in the reduced token representation derived from super-pixels, we introduce dynamic positional encodings anchored to the geometric centroids of each super-pixel region. For a given super-pixel $S_r$, the centroid coordinates $(c_x^r, c_y^r)$ are computed as the mean of the pixel coordinates contained within the region. These coordinates are subsequently normalized by the image dimensions $W$ and $H$, and projected into the embedding space $\mathbb{R}^D$ via a multilayer perceptron (MLP). The resulting positional encoding $PE_r$ is then added to the corresponding super patch embedding. This approach enables the model to retain spatial contextuality in a manner that aligns with the underlying image structure, as defined by the super-pixel segmentation, rather than relying on fixed-grid positional priors.

\begin{algorithm}[H]
\caption{Superpixel-based Patch Pooling (SPPP)}
\label{alg:sppp}
\begin{algorithmic}[1]
\Require Image $\mathcal{I}$, patch size $p \times p$, number of superpixels $K$
\Ensure Superpixel-level embeddings $\{h_i\}_{i=1}^K$
\State Partition $\mathcal{I}$ into fixed-size patches $\{P_j\}$
\State Apply SLIC to compute $K$ superpixels $\{S_i\}$
\For{each superpixel $S_i$}
    \State Compute overlap between $S_i$ and all patches $\{P_j\}$
    \State Aggregate patch features using overlap weights to form $h_i$
    \State Compute centroid $(c_{i,x}, c_{i,y})$ of $S_i$
    \State Normalize centroid with image dimensions
    \State Compute positional encoding via MLP
    \State Add positional encoding to $h_i$
\EndFor
\State \Return Super patch embeddings $\{h_i\}$
\end{algorithmic}
\end{algorithm}

\subsection {Light Latent Attention (LLA)}
\begin{figure}[hbt!]
    \centering
    \includegraphics[width=7cm]{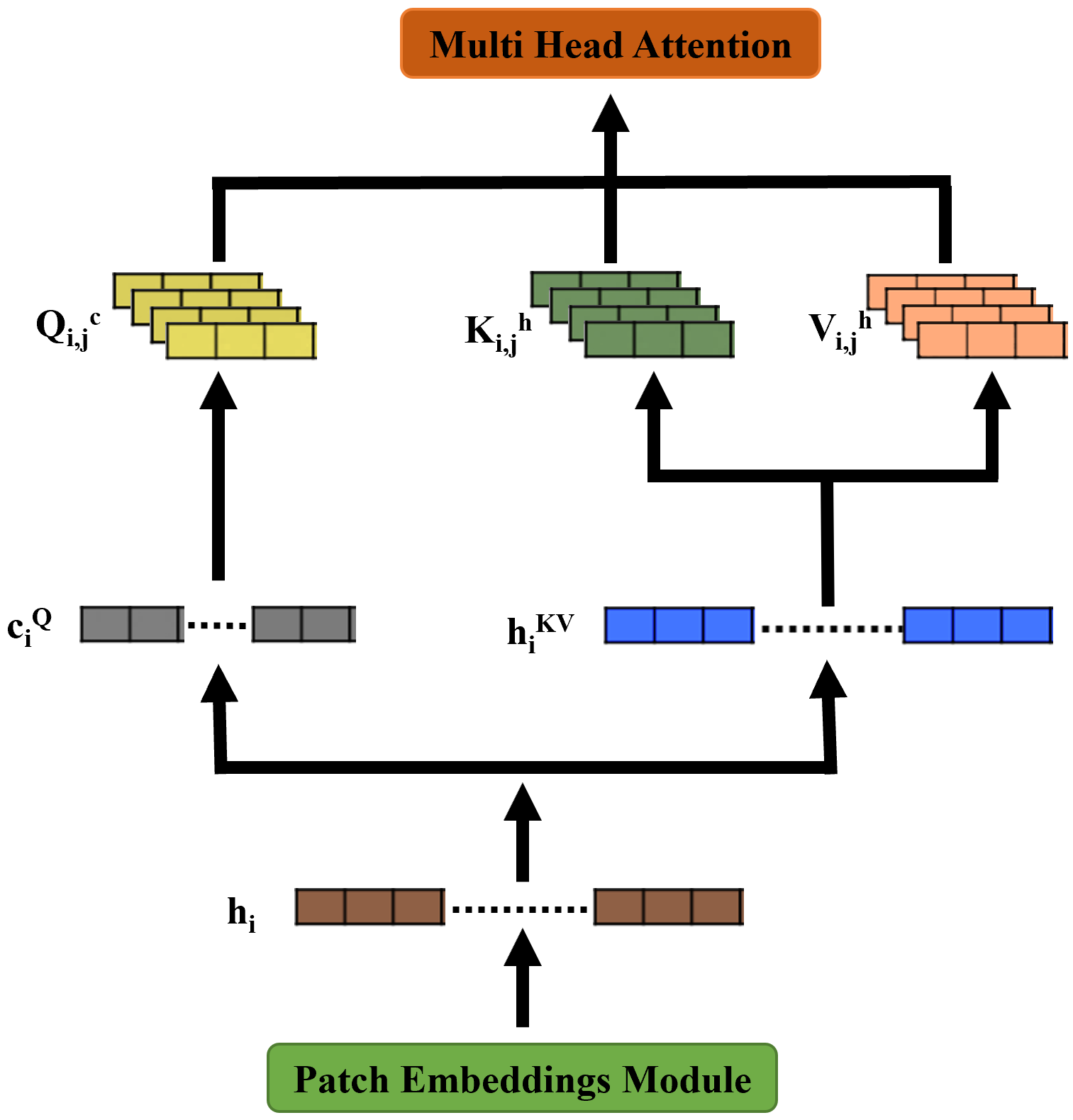}
    \caption{Architecture of the LLA module.}
    \label{fig4}
\end{figure}

We propose the LLA module to further reduce the dimensions of the embedding vectors using the concept of latent tokens. Figure\ref{fig4} illustrates the overall architecture of the LLA module.

As shown in the Figure\ref{fig4}, \(h_i \) are the patch embeddings vector, where \(h_i\in \mathbb{R}^{X\times D}\). Moving forward into the LLA module, the Key and Value vector \(h_i^{KV} \) is separately reshaped into appropriate Key and Value tensors based on the length of the input sequence, the number of attention heads and the dimensions of each head. On the other hand, we perform low rank compression of the Query vectors through learned projection matrices to generate latent representations \(c_t^{Q} \)\cite{deepseekai2024deepseekv2strongeconomicalefficient}, where  \(c_t^{Q} \in \mathbb{R}^{L \times D}\). It is worth noting that \((L << X\)). This compression is aimed to reduce memory consumption during model training. After that, the Query tensor derived from the latent representation and the Key and Value tensors derived from the original input are fed into the attention module to carry out cross-attention operations. The overall computations of the LLA module are summarized by following equations:
\begin{equation}
Q = c_t^{Q} W_Q \in \mathbb{R}^{B \times L \times D}
\end{equation}
\begin{equation}
K = h_i W_K \in \mathbb{R}^{B \times X \times D}, 
V= h_i W_V \in \mathbb{R}^{B \times X \times D}
\end{equation}
where \( \mathbf{W}_Q, \mathbf{W}_K, \mathbf{W}_V \in \mathbb{R}^{D \times D} \) are learned projection matrices, \(B\) is the batch size and \(L\), \(X\) are lengths of latent tokens and input sequences respectively. 

The K,Q and V are further reshaped to split them for feeding into multiple heads as follows:
\begin{equation}
\mathbf{Q} \rightarrow \mathbb{R}^{B \times H \times L \times d}, 
\mathbf{K}, \mathbf{V} \rightarrow \mathbb{R}^{B \times H \times T \times d}
\end{equation}
Where \( H \) is the number of heads such that \( d = D / H \). 
Finally we compute the scaled dot product cross-attention across the latent Query vectors and the input projected Key and Values vectors as follows:
\begin{equation}
\mathbf{A}_{b,h} = \mathrm{softmax} \left( \frac{\mathbf{Q}_{b,h} \mathbf{K}_{b,h}^\top}{\sqrt{d}} \right) \in \mathbb{R}^{L \times X}
\end{equation}
\begin{equation}
\mathbf{O}_{b,h} = \mathbf{A}_{b,h} \mathbf{V}_{b,h} \in \mathbb{R}^{L \times d}
\end{equation}
Outputs of the multiple heads are concatenated and progressed to the further layers of the model. 

\begin{algorithm}[H]
\caption{Light Latent Attention (LLA)}
\label{alg:lla}
\begin{algorithmic}[1]
\Require Superpixel embeddings $\{h_i\}$ from SPPP
\Ensure Attention-enhanced embeddings $\{z_i\}$
\For{each embedding $h_i$}
    \State Apply layer normalization to obtain $\tilde{h}_i$
    \State Split $\tilde{h}_i$ into latent query and key-value components
    \State Project components to form $Q_i$, $K_i$, and $V_i$
    \State Compute attention output using multi-head attention
\EndFor
\State \Return Computed Attentions 
\end{algorithmic}
\end{algorithm}

\section{Experimental Setup}

\subsection{Implementation Details}
Table\ref{tab1} shows different parameters comparing the base ViT architecture to our proposed Focused attention ViT. We intend to keep the architecture of our proposed model as close as possible to the original ViT while reducing the computational complexity of the attention mechanism through a combination of the SPPP and LLA modules. Regarding the training parameters, we consistently use the AdamW stochastic optimizer with a learning rate of $1 \times 10^{-4}$, batch size of $128$ and a weight decay of $0.05$ throughout our experiments. We train all the models for 50 epochs on respective datasets. For experiments involving pretrained weights, we use ImageNet weights trained using the original ViT model. All the experiments were carried out on a local machine using Intel Core i9 14900HX processor, NVIDIA RTX 4090 GPU (24 GB VRAM, $\sim$82 TFLOPS FP32) and 64 GB DDR4 RAM. For evaluation metrics, we use parameters such as accuracy($\%$), training time, inference speed(secs/image) and memory consumption(GBs) to assess the performance of all the models.   
\begin{table}
  \caption{Comparison of the Base ViT model with our proposed Focused attention ViT.}
  \label{sample-table}
  \centering
  \begin{tabular}{lcc}
    \toprule
    Component     & Base ViT     & Focused attention ViT \\
    \midrule
    Patch Size & $4\times4$ & $4\times4$     \\
    Embedding Dimension     & 768 & 768      \\
    Transformer Layers     & 12       & 12  \\
    Attention Heads     & 12       & 12  \\
    FFN Dimensions     & 768       & 768  \\
    No. of final patches     & N& S (where S<N)  \\
    Positional Encoding     & Learned Fixed Grid       & Learned Dynamic (Super-pixel Centroids)  \\
    Latent Token Dimensions     & ---       & L (where L<<N)  \\
    \bottomrule
  \end{tabular}
  \label{tab1}
\end{table}

\subsection{Datasets}
We use five different datasets to train our models for carrying our comparative analysis including CIFAR-10, CIFAR-100 \cite{articleKrizhevsky}, SVHN \cite{articleNetzer}, Fashion MNIST \cite{articleXiao} and STL10 \cite{pmlr-v15-coates11a}. Due to their low image dimensions, these datasets serve an ideal purpose to perform rapid experimental evaluation for the transformer-based architectures. The details of these datasets are included in the appendix.

\subsection{Results}

We train eight distinct models across all datasets, including the baseline Vision Transformer (ViT) initialized with both random and pre-trained weights. One variant, termed Cross-Attention ViT, employs a simplified cross-attention mechanism where the attention operates between compressed feature representations and the original patch embeddings. To enhance this further, the Multi-Head Cross-Attention ViT extends the mechanism across multiple attention heads, enabling the model to attend to diverse subspaces of the compressed features simultaneously. Finally, we evaluate our proposed Focused Attention ViT, which incorporates the SPPP and LLA modules to more effectively guide the attention process toward salient regions. The experimental results are summarized in Tables 3–6. The last three rows of each table represent the ablation experiments of our proposed Focused Attention ViT. Table\ref{tab3} reports the training times for all models. The standard ViT with random initialization exhibits the highest training times across all datasets. In contrast, the Cross-Attention and Multi-Head Cross-Attention ViTs achieve modest improvements in efficiency. The ViT model initialized with pre-trained weights converges significantly faster than its randomly initialized counterparts. Notably, our proposed Focused Attention ViT including SPPP and LLA modules achieves the greatest efficiency, reducing training times by more than 45\% across all datasets.

\begin{table}[h]
\centering
\caption{ Training Time (mins)}
\begin{tabular}{lccccc}
\toprule
Model & CIFAR-10 & CIFAR-100 & SVHN & FashionMNIST & STL-10 \\
\midrule
ViT & 186.67 & 202.85 & 275.00 & 225.00 & 78.33 \\
Pretrained ViT & 74.67 & 81.10 & 110.00 & 90.00 & 31.33 \\
Cross-Attention ViT & 171.67 & 186.33 & 251.67 & 206.67 & 72.33 \\
Multi-Head Cross ViT & 161.00 & 173.45 & 235.00 & 193.33 & 66.83 \\
ViT + SPPP & 97.67 & 106.92 & 143.17 & 117.17 & 39.00 \\
Pretrained ViT + SPPP & 39.00 & 42.80 & 57.33 & 46.83 & 15.67 \\
Pretrained ViT + LLA & 50.70 & 55.98 & 74.53 & 60.88 & 17.23 \\
\textbf{Pretrained ViT} + SPPP + LLA & \textbf{38.42} & \textbf{41.77} & \textbf{56.85} & \textbf{46.29} & \textbf{15.34} \\
\bottomrule
\end{tabular}
\label{tab3}
\end{table}

Table\ref{tab4} shows the performance of all models in terms of the percentage accuracy achieved. The pre-trained ViT achieves the best accuracy values overall. Focused attention ViT also achieves comparative accuracy values but with significantly lower training times as mentioned in Table\ref{tab1}.  
\begin{table}[h]
\centering
\caption{Accuracy (\%) \small (All results within $\pm$0.5 threshold)}
\begin{tabular}{lccccc}
\toprule
Model & CIFAR10 & CIFAR100 & SVHN & FashionMNIST & STL10 \\
\midrule
ViT & 95.4 & 81.2 & 96.7 & 92.3 & 90.5 \\
\textbf{Pretrained ViT} & \textbf{96.7} & \textbf{82.3} & \textbf{97.8} & \textbf{93.3} & \textbf{91.6} \\
Cross-Attention ViT & 95.6 & 80.7 & 96.5 & 92.2 & 90.4 \\
Multi-Head Cross ViT & 95.8 & 81.0 & 96.8 & 92.4 & 90.6 \\
ViT + SPPP & 94.3 & 79.0 & 95.5 & 91.0 & 89.1 \\
Pretrained ViT + SPPP& 95.7 & 80.3 & 96.6 & 92.1 & 90.3 \\
Pretrained ViT + LLA & 96.0 & 81.2 & 96.9 & 92.6 & 90.8 \\
Pretrained ViT + SPPP + LLA & 95.9 & 80.9 & 96.7 & 92.5 & 90.7 \\
\bottomrule
\end{tabular}
\label{tab4}
\end{table}

Similarly, Table\ref{tab5} shows the memory consumption of each model with respect to different datasets. The proposed Focused attention ViT has approximately 90\% less memory consumption as compared to the other models signifying its light-weight abilities. Finally, Table\ref{tab6} shows the inference times taken per image for all the datasets included in the study. It can be seen that due to the light-weight architecture of our newly proposed model, it has the fastest inference times compared to other models as well. 

\begin{table}[h]
\centering
\caption{Memory Usage (GBs) during Training}
\begin{tabular}{lccccc}
\toprule
Model & CIFAR-10 & CIFAR-100 & SVHN & FashionMNIST & STL-10 \\
\midrule
ViT & 5.3 & 5.6 & 5.2 & 5.1 & 17.4 \\
Pretrained ViT & 5.1 & 5.4 & 5.0 & 5.0 & 16.8 \\
Cross-Attention ViT & 4.2 & 4.5 & 4.1 & 4.0 & 13.5 \\
Multi-Head Cross ViT & 4.6 & 4.9 & 4.4 & 4.3 & 14.2 \\
ViT + SPPP & 0.42 & 0.45 & 0.40 & 0.39 & 0.72 \\
Pretrained ViT + SPPP& 0.41 & 0.44 & 0.39 & 0.38 & 0.71 \\
Pretrained ViT + LLA & 0.50 & 0.54 & 0.47 & 0.46 & 0.85 \\
\textbf{Pretrained ViT + SPPP + LLA} & \textbf{0.39} & \textbf{0.43} & \textbf{0.37} & \textbf{0.36} & \textbf{0.67} \\
\bottomrule
\end{tabular}
\label{tab5}
\end{table}

\begin{table}[h]
\centering
\caption{Inference Time (secs/image)}
\begin{tabular}{lccccc}
\toprule
Model & CIFAR-10 & CIFAR-100 & SVHN & FashionMNIST & STL-10 \\
\midrule
ViT & 0.102 & 0.106 & 0.098 & 0.094 & 1.12 \\
Pretrained ViT & 0.097 & 0.102 & 0.092 & 0.089 & 1.08 \\
Cross-Attention ViT & 0.090 & 0.095 & 0.088 & 0.085 & 1.00 \\
Multi-Head Cross ViT & 0.083 & 0.089 & 0.081 & 0.078 & 0.95 \\
ViT + SPPP & 0.048 & 0.052 & 0.046 & 0.043 & 0.21 \\
Pretrained ViT + SPPP & 0.045 & 0.050 & 0.043 & 0.041 & 0.19 \\
Pretrained ViT + LLA & 0.064 & 0.069 & 0.061 & 0.059 & 0.36 \\
\textbf{Pretrained ViT + SPPP + LLA} & \textbf{0.043} & \textbf{0.047} & \textbf{0.041} & \textbf{0.039} & \textbf{0.18} \\
\bottomrule
\end{tabular}
\label{tab6}
\end{table}
\vspace{-0.3cm}

\section{Discussion and Conclusion}
As demonstrated by our experimental results, our proposed Focused Attention ViT significantly improves the computational efficiency including training time, memory usage and inference speed while maintaining comparable accuracy to the original ViT. These improvements are achieved by incorporating the SPPP and LLA modules in the ViT. The SPPP module implements the strategy of adaptive spatial pooling of the patches based on super-pixel clustering to generate semantically meaningful super patches. This approach yields rich patch embedding with less number of embedding vectors compared to the original ViT. Effectively, the SPPP module reduces the space complexity of the ViT architecture from \(\mathcal{O}(N^2)\) to \(\mathcal{O}(S^2)\), where \(S < N\). The LLA module further enhances the efficiency by reducing the set of Query vectors through intermediate learnable latent tokens before passing them to the attention heads. The LLA reduces the complexity \(\mathcal{O}(S^2)\) to \(\mathcal{O}(L \times S)\), where \(L << S\). Therefore, both modules play their part in effectively reducing the embedding tokens while retaining useful feature information necessary to train the model. This results in a considerable reduction in the computational overhead of the attention mechanism making the overall ViT architecture more robust and light-weight.     

In conclusion, the computationally intensive nature of the self-attention mechanism poses significant challenges for deploying ViT-based architectures on high-resolution images under resource-constrained settings. Similarly, training convergence for ViTs is also difficult due to the requirement of huge training datasets. To address these issues, we proposed the Focused Attention ViT by introducing two novel modules called SPPP and LLA. These modules collectively improve the computational performance of the ViT architecture by leveraging the inherent semantic structure of the input images and generating learnable latent token representations respectively. Our results demonstrate substantial improvements in model training times, inference speed and memory efficiency while achieving comparable accuracy with the standard ViT model. This work is a continuation of the efforts toward the development of efficient and robust ViT-based architectures that have the possibility of being implemented on edge devices under real-time settings. We are excited about the future prospects of this work in developing resource-friendly and data-efficient models. In future, we plan to explore the possibility and impact of using a latent-heavy implementation of the attention mechanism in the ViT architecture as well as testing our results on more complex datasets with larger image dimensions.

\bibliographystyle{plainnat}

\bibliography{References.bib}

\begin{thebibliography}{52}
\providecommand{\natexlab}[1]{#1}
\providecommand{\url}[1]{\texttt{#1}}
\expandafter\ifx\csname urlstyle\endcsname\relax
  \providecommand{\doi}[1]{doi: #1}\else
  \providecommand{\doi}{doi: \begingroup \urlstyle{rm}\Url}\fi

\bibitem[Aburass et~al.(2025)Aburass, Dorgham, Al~Shaqsi, Abu~Rumman, and Al-Kadi]{Aburass2025}
Sanad Aburass, Osama Dorgham, Jamil Al~Shaqsi, Maha Abu~Rumman, and Omar Al-Kadi.
\newblock Vision transformers in medical imaging: a comprehensive review of advancements and applications across multiple diseases.
\newblock \emph{Journal of Imaging Informatics in Medicine}, Mar 2025.
\newblock ISSN 2948-2933.
\newblock \doi{10.1007/s10278-025-01481-y}.
\newblock URL \url{https://doi.org/10.1007/s10278-025-01481-y}.

\bibitem[Achanta et~al.(2012)Achanta, Shaji, Smith, Lucchi, Fua, and Süsstrunk]{6205760}
Radhakrishna Achanta, Appu Shaji, Kevin Smith, Aurelien Lucchi, Pascal Fua, and Sabine Süsstrunk.
\newblock Slic superpixels compared to state-of-the-art superpixel methods.
\newblock \emph{IEEE Transactions on Pattern Analysis and Machine Intelligence}, 34\penalty0 (11):\penalty0 2274--2282, 2012.
\newblock \doi{10.1109/TPAMI.2012.120}.

\bibitem[Ahn et~al.(2023)Ahn, Kim, Hong, and Chul~Ko]{10030830}
Dasom Ahn, Sangwon Kim, Hyunsu Hong, and Byoung Chul~Ko.
\newblock Star-transformer: A spatio-temporal cross attention transformer for human action recognition.
\newblock In \emph{2023 IEEE/CVF Winter Conference on Applications of Computer Vision (WACV)}, pages 3319--3328, 2023.
\newblock \doi{10.1109/WACV56688.2023.00333}.

\bibitem[Ambartsoumian and Popowich(2018)]{ambartsoumian-popowich-2018-self}
Artaches Ambartsoumian and Fred Popowich.
\newblock Self-attention: A better building block for sentiment analysis neural network classifiers.
\newblock In Alexandra Balahur, Saif~M. Mohammad, Veronique Hoste, and Roman Klinger, editors, \emph{Proceedings of the 9th Workshop on Computational Approaches to Subjectivity, Sentiment and Social Media Analysis}, pages 130--139, Brussels, Belgium, October 2018. Association for Computational Linguistics.
\newblock \doi{10.18653/v1/W18-6219}.
\newblock URL \url{https://aclanthology.org/W18-6219/}.

\bibitem[Arnab et~al.(2021)Arnab, Dehghani, Heigold, Sun, Lučić, and Schmid]{9710415}
Anurag Arnab, Mostafa Dehghani, Georg Heigold, Chen Sun, Mario Lučić, and Cordelia Schmid.
\newblock Vivit: A video vision transformer.
\newblock In \emph{2021 IEEE/CVF International Conference on Computer Vision (ICCV)}, pages 6816--6826, 2021.
\newblock \doi{10.1109/ICCV48922.2021.00676}.

\bibitem[Bertasius et~al.(2021)Bertasius, Wang, and Torresani]{pmlr-v139-bertasius21a}
Gedas Bertasius, Heng Wang, and Lorenzo Torresani.
\newblock Is space-time attention all you need for video understanding?
\newblock In Marina Meila and Tong Zhang, editors, \emph{Proceedings of the 38th International Conference on Machine Learning}, volume 139 of \emph{Proceedings of Machine Learning Research}, pages 813--824. PMLR, 18--24 Jul 2021.
\newblock URL \url{https://proceedings.mlr.press/v139/bertasius21a.html}.

\bibitem[Carion et~al.(2020)Carion, Massa, Synnaeve, Usunier, Kirillov, and Zagoruyko]{10.1007/978-3-030-58452-8_13}
Nicolas Carion, Francisco Massa, Gabriel Synnaeve, Nicolas Usunier, Alexander Kirillov, and Sergey Zagoruyko.
\newblock End-to-end object detection with transformers.
\newblock In \emph{Computer Vision – ECCV 2020: 16th European Conference, Glasgow, UK, August 23–28, 2020, Proceedings, Part I}, page 213–229, Berlin, Heidelberg, 2020. Springer-Verlag.
\newblock ISBN 978-3-030-58451-1.
\newblock \doi{10.1007/978-3-030-58452-8_13}.
\newblock URL \url{https://doi.org/10.1007/978-3-030-58452-8_13}.

\bibitem[Chen et~al.(2021)Chen, Fan, and Panda]{9711309}
Chun-Fu~Richard Chen, Quanfu Fan, and Rameswar Panda.
\newblock Crossvit: Cross-attention multi-scale vision transformer for image classification.
\newblock In \emph{2021 IEEE/CVF International Conference on Computer Vision (ICCV)}, pages 347--356, 2021.
\newblock \doi{10.1109/ICCV48922.2021.00041}.

\bibitem[Chen et~al.(2024)Chen, Wu, Chen, Wu, Zhang, and Li]{article123}
Dong Chen, Peisong Wu, Mingdong Chen, Mengtao Wu, Tao Zhang, and Chuanqi Li.
\newblock Ls-vit: Vision transformer for action recognition based on long and short-term temporal difference.
\newblock \emph{Frontiers in Neurorobotics}, 18, 10 2024.
\newblock \doi{10.3389/fnbot.2024.1457843}.

\bibitem[Coates et~al.(2011)Coates, Ng, and Lee]{pmlr-v15-coates11a}
Adam Coates, Andrew Ng, and Honglak Lee.
\newblock An analysis of single-layer networks in unsupervised feature learning.
\newblock In Geoffrey Gordon, David Dunson, and Miroslav Dudík, editors, \emph{Proceedings of the Fourteenth International Conference on Artificial Intelligence and Statistics}, volume~15 of \emph{Proceedings of Machine Learning Research}, pages 215--223, Fort Lauderdale, FL, USA, 11--13 Apr 2011. PMLR.
\newblock URL \url{https://proceedings.mlr.press/v15/coates11a.html}.

\bibitem[Devlin et~al.(2018)Devlin, Chang, Lee, and Toutanova]{DBLP:journals/corr/abs-1810-04805}
Jacob Devlin, Ming{-}Wei Chang, Kenton Lee, and Kristina Toutanova.
\newblock {BERT:} pre-training of deep bidirectional transformers for language understanding.
\newblock \emph{CoRR}, abs/1810.04805, 2018.
\newblock URL \url{http://arxiv.org/abs/1810.04805}.

\bibitem[Dosovitskiy et~al.(2021)Dosovitskiy, Beyer, Kolesnikov, Weissenborn, Zhai, Unterthiner, Dehghani, Minderer, Heigold, Gelly, Uszkoreit, and Houlsby]{DBLP:conf/iclr/DosovitskiyB0WZ21}
Alexey Dosovitskiy, Lucas Beyer, Alexander Kolesnikov, Dirk Weissenborn, Xiaohua Zhai, Thomas Unterthiner, Mostafa Dehghani, Matthias Minderer, Georg Heigold, Sylvain Gelly, Jakob Uszkoreit, and Neil Houlsby.
\newblock An image is worth 16x16 words: Transformers for image recognition at scale.
\newblock In \emph{9th International Conference on Learning Representations, {ICLR} 2021, Virtual Event, Austria, May 3-7, 2021}. OpenReview.net, 2021.
\newblock URL \url{https://openreview.net/forum?id=YicbFdNTTy}.

\bibitem[et. al.(2024)]{deepseekai2024deepseekv2strongeconomicalefficient}
Aixin~Liu et. al.
\newblock Deepseek-v2: A strong, economical, and efficient mixture-of-experts language model, 2024.
\newblock URL \url{https://arxiv.org/abs/2405.04434}.

\bibitem[et. al.(2025)]{deepseekai2025deepseekv3technicalreport}
Aixin~Liu et. al.
\newblock Deepseek-v3 technical report, 2025.
\newblock URL \url{https://arxiv.org/abs/2412.19437}.

\bibitem[Fuller et~al.(2024)Fuller, Kyrollos, Yassin, and Green]{fuller2024lookhere}
Anthony Fuller, Daniel Kyrollos, Yousef Yassin, and James~R Green.
\newblock Lookhere: Vision transformers with directed attention generalize and extrapolate.
\newblock In \emph{The Thirty-eighth Annual Conference on Neural Information Processing Systems}, 2024.
\newblock URL \url{https://openreview.net/forum?id=o7DOGbZeyP}.

\bibitem[Graham et~al.(2021)Graham, El-Nouby, Touvron, Stock, Joulin, Jégou, and Douze]{9711161}
Ben Graham, Alaaeldin El-Nouby, Hugo Touvron, Pierre Stock, Armand Joulin, Hervé Jégou, and Matthijs Douze.
\newblock Levit: a vision transformer in convnet’s clothing for faster inference.
\newblock In \emph{2021 IEEE/CVF International Conference on Computer Vision (ICCV)}, pages 12239--12249, 2021.
\newblock \doi{10.1109/ICCV48922.2021.01204}.

\bibitem[Hatamizadeh et~al.(2022)Hatamizadeh, Tang, Nath, Yang, Myronenko, Landman, Roth, and Xu]{9706678}
Ali Hatamizadeh, Yucheng Tang, Vishwesh Nath, Dong Yang, Andriy Myronenko, Bennett Landman, Holger~R. Roth, and Daguang Xu.
\newblock Unetr: Transformers for 3d medical image segmentation.
\newblock In \emph{2022 IEEE/CVF Winter Conference on Applications of Computer Vision (WACV)}, pages 1748--1758, 2022.
\newblock \doi{10.1109/WACV51458.2022.00181}.

\bibitem[He et~al.(2023)He, Gan, Li, Rekik, Yin, Ji, Gao, Wang, Zhang, and Shen]{HE202359}
Kelei He, Chen Gan, Zhuoyuan Li, Islem Rekik, Zihao Yin, Wen Ji, Yang Gao, Qian Wang, Junfeng Zhang, and Dinggang Shen.
\newblock Transformers in medical image analysis.
\newblock \emph{Intelligent Medicine}, 3\penalty0 (1):\penalty0 59--78, 2023.
\newblock ISSN 2667-1026.
\newblock \doi{https://doi.org/10.1016/j.imed.2022.07.002}.
\newblock URL \url{https://www.sciencedirect.com/science/article/pii/S2667102622000717}.

\bibitem[Heo et~al.(2021)Heo, Yun, Han, Chun, Choe, and Oh]{9710548}
Byeongho Heo, Sangdoo Yun, Dongyoon Han, Sanghyuk Chun, Junsuk Choe, and Seong~Joon Oh.
\newblock Rethinking spatial dimensions of vision transformers.
\newblock In \emph{2021 IEEE/CVF International Conference on Computer Vision (ICCV)}, pages 11916--11925, 2021.
\newblock \doi{10.1109/ICCV48922.2021.01172}.

\bibitem[Hong et~al.(2022)Hong, Han, Yao, Gao, Zhang, Plaza, and Chanussot]{9627165}
Danfeng Hong, Zhu Han, Jing Yao, Lianru Gao, Bing Zhang, Antonio Plaza, and Jocelyn Chanussot.
\newblock Spectralformer: Rethinking hyperspectral image classification with transformers.
\newblock \emph{IEEE Transactions on Geoscience and Remote Sensing}, 60:\penalty0 1--15, 2022.
\newblock \doi{10.1109/TGRS.2021.3130716}.

\bibitem[Krizhevsky(2012)]{articleKrizhevsky}
Alex Krizhevsky.
\newblock Learning multiple layers of features from tiny images.
\newblock \emph{University of Toronto}, 05 2012.

\bibitem[Li et~al.(2023)Li, Liu, Ding, Liu, Wang, and Yang]{9674785}
Wenhao Li, Hong Liu, Runwei Ding, Mengyuan Liu, Pichao Wang, and Wenming Yang.
\newblock Exploiting temporal contexts with strided transformer for 3d human pose estimation.
\newblock \emph{IEEE Transactions on Multimedia}, 25:\penalty0 1282--1293, 2023.
\newblock \doi{10.1109/TMM.2022.3141231}.

\bibitem[Lin et~al.(2017)Lin, Feng, dos Santos, Yu, Xiang, Zhou, and Bengio]{DBLP:journals/corr/LinFSYXZB17}
Zhouhan Lin, Minwei Feng, C{\'{\i}}cero~Nogueira dos Santos, Mo~Yu, Bing Xiang, Bowen Zhou, and Yoshua Bengio.
\newblock A structured self-attentive sentence embedding.
\newblock \emph{CoRR}, abs/1703.03130, 2017.
\newblock URL \url{http://arxiv.org/abs/1703.03130}.

\bibitem[Liu et~al.(2022{\natexlab{a}})Liu, Li, Zhang, Yang, Qi, Su, Zhu, and Zhang]{liu2022dabdetr}
Shilong Liu, Feng Li, Hao Zhang, Xiao Yang, Xianbiao Qi, Hang Su, Jun Zhu, and Lei Zhang.
\newblock {DAB}-{DETR}: Dynamic anchor boxes are better queries for {DETR}.
\newblock In \emph{International Conference on Learning Representations}, 2022{\natexlab{a}}.
\newblock URL \url{https://openreview.net/forum?id=oMI9PjOb9Jl}.

\bibitem[Liu et~al.(2021)Liu, Lin, Cao, Hu, Wei, Zhang, Lin, and Guo]{9710580}
Ze~Liu, Yutong Lin, Yue Cao, Han Hu, Yixuan Wei, Zheng Zhang, Stephen Lin, and Baining Guo.
\newblock Swin transformer: Hierarchical vision transformer using shifted windows.
\newblock In \emph{2021 IEEE/CVF International Conference on Computer Vision (ICCV)}, pages 9992--10002, 2021.
\newblock \doi{10.1109/ICCV48922.2021.00986}.

\bibitem[Liu et~al.(2022{\natexlab{b}})Liu, Hu, Lin, Yao, Xie, Wei, Ning, Cao, Zhang, Dong, Wei, and Guo]{9879380}
Ze~Liu, Han Hu, Yutong Lin, Zhuliang Yao, Zhenda Xie, Yixuan Wei, Jia Ning, Yue Cao, Zheng Zhang, Li~Dong, Furu Wei, and Baining Guo.
\newblock Swin transformer v2: Scaling up capacity and resolution.
\newblock In \emph{2022 IEEE/CVF Conference on Computer Vision and Pattern Recognition (CVPR)}, pages 11999--12009, 2022{\natexlab{b}}.
\newblock \doi{10.1109/CVPR52688.2022.01170}.

\bibitem[Mehta and Rastegari(2022)]{mehta2022mobilevit}
Sachin Mehta and Mohammad Rastegari.
\newblock Mobilevit: Light-weight, general-purpose, and mobile-friendly vision transformer.
\newblock In \emph{International Conference on Learning Representations}, 2022.
\newblock URL \url{https://openreview.net/forum?id=vh-0sUt8HlG}.

\bibitem[Meng et~al.(2022)Meng, Li, Chen, Lan, Wu, Jiang, and Lim]{9879366}
Lingchen Meng, Hengduo Li, Bor-Chun Chen, Shiyi Lan, Zuxuan Wu, Yu-Gang Jiang, and Ser-Nam Lim.
\newblock Adavit: Adaptive vision transformers for efficient image recognition.
\newblock In \emph{2022 IEEE/CVF Conference on Computer Vision and Pattern Recognition (CVPR)}, pages 12299--12308, 2022.
\newblock \doi{10.1109/CVPR52688.2022.01199}.

\bibitem[Netzer et~al.(2011)Netzer, Wang, Coates, Bissacco, Wu, and Ng]{articleNetzer}
Yuval Netzer, Tao Wang, Adam Coates, Alessandro Bissacco, Bo~Wu, and Andrew Ng.
\newblock Reading digits in natural images with unsupervised feature learning.
\newblock \emph{NIPS}, 01 2011.

\bibitem[Radford and Narasimhan(2018)]{Radford2018ImprovingLU}
Alec Radford and Karthik Narasimhan.
\newblock Improving language understanding by generative pre-training.
\newblock 2018.
\newblock URL \url{https://api.semanticscholar.org/CorpusID:49313245}.

\bibitem[Rao et~al.(2021)Rao, Zhao, Liu, Lu, Zhou, and Hsieh]{NEURIPS2021_747d3443}
Yongming Rao, Wenliang Zhao, Benlin Liu, Jiwen Lu, Jie Zhou, and Cho-Jui Hsieh.
\newblock Dynamicvit: Efficient vision transformers with dynamic token sparsification.
\newblock In M.~Ranzato, A.~Beygelzimer, Y.~Dauphin, P.S. Liang, and J.~Wortman Vaughan, editors, \emph{Advances in Neural Information Processing Systems}, volume~34, pages 13937--13949. Curran Associates, Inc., 2021.
\newblock URL \url{https://proceedings.neurips.cc/paper_files/paper/2021/file/747d3443e319a22747fbb873e8b2f9f2-Paper.pdf}.

\bibitem[Ryoo et~al.(2021)Ryoo, Piergiovanni, Arnab, Dehghani, and Angelova]{NEURIPS2021_6a30e32e}
Michael Ryoo, AJ~Piergiovanni, Anurag Arnab, Mostafa Dehghani, and Anelia Angelova.
\newblock Tokenlearner: Adaptive space-time tokenization for videos.
\newblock In M.~Ranzato, A.~Beygelzimer, Y.~Dauphin, P.S. Liang, and J.~Wortman Vaughan, editors, \emph{Advances in Neural Information Processing Systems}, volume~34, pages 12786--12797. Curran Associates, Inc., 2021.
\newblock URL \url{https://proceedings.neurips.cc/paper_files/paper/2021/file/6a30e32e56fce5cf381895dfe6ca7b6f-Paper.pdf}.

\bibitem[Setyawan et~al.(2025)Setyawan, Sun, Hsu, Kuo, and Hsieh]{setyawan2025microvitvisiontransformerlow}
Novendra Setyawan, Chi-Chia Sun, Mao-Hsiu Hsu, Wen-Kai Kuo, and Jun-Wei Hsieh.
\newblock Microvit: A vision transformer with low complexity self attention for edge device, 2025.
\newblock URL \url{https://arxiv.org/abs/2502.05800}.

\bibitem[Thisanke et~al.(2023)Thisanke, Deshan, Chamith, Seneviratne, Vidanaarachchi, and Herath]{THISANKE2023106669}
Hans Thisanke, Chamli Deshan, Kavindu Chamith, Sachith Seneviratne, Rajith Vidanaarachchi, and Damayanthi Herath.
\newblock Semantic segmentation using vision transformers: A survey.
\newblock \emph{Engineering Applications of Artificial Intelligence}, 126:\penalty0 106669, 2023.
\newblock ISSN 0952-1976.
\newblock \doi{https://doi.org/10.1016/j.engappai.2023.106669}.
\newblock URL \url{https://www.sciencedirect.com/science/article/pii/S0952197623008539}.

\bibitem[Touvron et~al.(2021)Touvron, Cord, Douze, Massa, Sablayrolles, and Jegou]{pmlr-v139-touvron21a}
Hugo Touvron, Matthieu Cord, Matthijs Douze, Francisco Massa, Alexandre Sablayrolles, and Herve Jegou.
\newblock Training data-efficient image transformers \&amp; distillation through attention.
\newblock In Marina Meila and Tong Zhang, editors, \emph{Proceedings of the 38th International Conference on Machine Learning}, volume 139 of \emph{Proceedings of Machine Learning Research}, pages 10347--10357. PMLR, 18--24 Jul 2021.
\newblock URL \url{https://proceedings.mlr.press/v139/touvron21a.html}.

\bibitem[Vaswani et~al.(2017)Vaswani, Shazeer, Parmar, Uszkoreit, Jones, Gomez, Kaiser, and Polosukhin]{NIPS2017_3f5ee243}
Ashish Vaswani, Noam Shazeer, Niki Parmar, Jakob Uszkoreit, Llion Jones, Aidan~N Gomez, \L~ukasz Kaiser, and Illia Polosukhin.
\newblock Attention is all you need.
\newblock In \emph{Advances in Neural Information Processing Systems}, volume~30, 2017.

\bibitem[Wang et~al.(2020)Wang, Li, Khabsa, Fang, and Ma]{DBLP:journals/corr/abs-2006-04768}
Sinong Wang, Belinda~Z. Li, Madian Khabsa, Han Fang, and Hao Ma.
\newblock Linformer: Self-attention with linear complexity.
\newblock \emph{CoRR}, abs/2006.04768, 2020.
\newblock URL \url{https://arxiv.org/abs/2006.04768}.

\bibitem[Wang et~al.(2025)Wang, Deng, Zheng, Chattopadhyay, and Wang]{technologies13010032}
Yaoli Wang, Yaojun Deng, Yuanjin Zheng, Pratik Chattopadhyay, and Lipo Wang.
\newblock Vision transformers for image classification: A comparative survey.
\newblock \emph{Technologies}, 13\penalty0 (1), 2025.
\newblock ISSN 2227-7080.
\newblock \doi{10.3390/technologies13010032}.
\newblock URL \url{https://www.mdpi.com/2227-7080/13/1/32}.

\bibitem[Xiao et~al.(2017)Xiao, Rasul, and Vollgraf]{articleXiao}
Han Xiao, Kashif Rasul, and Roland Vollgraf.
\newblock Fashion-mnist: a novel image dataset for benchmarking machine learning algorithms.
\newblock 08 2017.
\newblock \doi{10.48550/arXiv.1708.07747}.

\bibitem[Xie et~al.(2021)Xie, Wang, Yu, Anandkumar, Alvarez, and Luo]{NEURIPS2021_64f1f27b}
Enze Xie, Wenhai Wang, Zhiding Yu, Anima Anandkumar, Jose~M. Alvarez, and Ping Luo.
\newblock Segformer: Simple and efficient design for semantic segmentation with transformers.
\newblock In M.~Ranzato, A.~Beygelzimer, Y.~Dauphin, P.S. Liang, and J.~Wortman Vaughan, editors, \emph{Advances in Neural Information Processing Systems}, volume~34, pages 12077--12090. Curran Associates, Inc., 2021.
\newblock URL \url{https://proceedings.neurips.cc/paper_files/paper/2021/file/64f1f27bf1b4ec22924fd0acb550c235-Paper.pdf}.

\bibitem[Xiong et~al.(2021)Xiong, Zeng, Chakraborty, Tan, Fung, Li, and Singh]{Xiong_Zeng_Chakraborty_Tan_Fung_Li_Singh_2021}
Yunyang Xiong, Zhanpeng Zeng, Rudrasis Chakraborty, Mingxing Tan, Glenn Fung, Yin Li, and Vikas Singh.
\newblock Nyströmformer: A nyström-based algorithm for approximating self-attention.
\newblock \emph{Proceedings of the AAAI Conference on Artificial Intelligence}, 35\penalty0 (16):\penalty0 14138--14148, May 2021.
\newblock \doi{10.1609/aaai.v35i16.17664}.
\newblock URL \url{https://ojs.aaai.org/index.php/AAAI/article/view/17664}.

\bibitem[Yang et~al.(2022)Yang, Wang, Zhang, Zhang, Wei, Lin, and Yuille]{9879515}
Chenglin Yang, Yilin Wang, Jianming Zhang, He~Zhang, Zijun Wei, Zhe Lin, and Alan Yuille.
\newblock Lite vision transformer with enhanced self-attention.
\newblock In \emph{2022 IEEE/CVF Conference on Computer Vision and Pattern Recognition (CVPR)}, pages 11988--11998, 2022.
\newblock \doi{10.1109/CVPR52688.2022.01169}.

\bibitem[Yang et~al.(2019)Yang, Dai, Yang, Carbonell, Salakhutdinov, and Le]{NEURIPS2019_dc6a7e65}
Zhilin Yang, Zihang Dai, Yiming Yang, Jaime Carbonell, Russ~R Salakhutdinov, and Quoc~V Le.
\newblock Xlnet: Generalized autoregressive pretraining for language understanding.
\newblock In H.~Wallach, H.~Larochelle, A.~Beygelzimer, F.~d\textquotesingle Alch\'{e}-Buc, E.~Fox, and R.~Garnett, editors, \emph{Advances in Neural Information Processing Systems}, volume~32. Curran Associates, Inc., 2019.
\newblock URL \url{https://proceedings.neurips.cc/paper_files/paper/2019/file/dc6a7e655d7e5840e66733e9ee67cc69-Paper.pdf}.

\bibitem[Yao et~al.(2023)Yao, Zhang, Li, Hong, and Chanussot]{10147258}
Jing Yao, Bing Zhang, Chenyu Li, Danfeng Hong, and Jocelyn Chanussot.
\newblock Extended vision transformer (exvit) for land use and land cover classification: A multimodal deep learning framework.
\newblock \emph{IEEE Transactions on Geoscience and Remote Sensing}, 61:\penalty0 1--15, 2023.
\newblock \doi{10.1109/TGRS.2023.3284671}.

\bibitem[You et~al.(2023)You, Shi, Guo, and Lin]{NEURIPS202369c49f75}
Haoran You, Huihong Shi, Yipin Guo, and Yingyan Lin.
\newblock Shiftaddvit: Mixture of multiplication primitives towards efficient vision transformer.
\newblock In A.~Oh, T.~Naumann, A.~Globerson, K.~Saenko, M.~Hardt, and S.~Levine, editors, \emph{Advances in Neural Information Processing Systems}, volume~36, pages 33319--33337. Curran Associates, Inc., 2023.
\newblock URL \url{https://proceedings.neurips.cc/paper_files/paper/2023/file/69c49f75ca31620f1f0d38093d9f3d9b-Paper-Conference.pdf}.

\bibitem[Yu et~al.(2022)Yu, Luo, Zhou, Si, Zhou, Wang, Feng, and Yan]{9879612}
Weihao Yu, Mi~Luo, Pan Zhou, Chenyang Si, Yichen Zhou, Xinchao Wang, Jiashi Feng, and Shuicheng Yan.
\newblock Metaformer is actually what you need for vision.
\newblock In \emph{2022 IEEE/CVF Conference on Computer Vision and Pattern Recognition (CVPR)}, pages 10809--10819, 2022.
\newblock \doi{10.1109/CVPR52688.2022.01055}.

\bibitem[Yuan et~al.(2021)Yuan, Chen, Wang, Yu, Shi, Jiang, Tay, Feng, and Yan]{9710747}
Li~Yuan, Yunpeng Chen, Tao Wang, Weihao Yu, Yujun Shi, Zihang Jiang, Francis E.~H. Tay, Jiashi Feng, and Shuicheng Yan.
\newblock Tokens-to-token vit: Training vision transformers from scratch on imagenet.
\newblock In \emph{2021 IEEE/CVF International Conference on Computer Vision (ICCV)}, pages 538--547, 2021.
\newblock \doi{10.1109/ICCV48922.2021.00060}.

\bibitem[YUAN et~al.(2021)YUAN, Fu, Huang, Lin, Zhang, Chen, and Wang]{NEURIPS2021_3bbfdde8}
YUHUI YUAN, Rao Fu, Lang Huang, Weihong Lin, Chao Zhang, Xilin Chen, and Jingdong Wang.
\newblock Hrformer: High-resolution vision transformer for dense predict.
\newblock In M.~Ranzato, A.~Beygelzimer, Y.~Dauphin, P.S. Liang, and J.~Wortman Vaughan, editors, \emph{Advances in Neural Information Processing Systems}, volume~34, pages 7281--7293. Curran Associates, Inc., 2021.
\newblock URL \url{https://proceedings.neurips.cc/paper_files/paper/2021/file/3bbfdde8842a5c44a0323518eec97cbe-Paper.pdf}.

\bibitem[Zhang et~al.(2025)Zhang, Yue, Fu, and Wu]{Zhang2025}
Chong Zhang, Jie Yue, Jianglong Fu, and Shouluan Wu.
\newblock River floating object detection with transformer model in real time.
\newblock \emph{Scientific Reports}, 15\penalty0 (1):\penalty0 9026, Mar 2025.
\newblock ISSN 2045-2322.
\newblock \doi{10.1038/s41598-025-93659-1}.
\newblock URL \url{https://doi.org/10.1038/s41598-025-93659-1}.

\bibitem[Zhao et~al.(2024)Zhao, Lv, Xu, Wei, Wang, Dang, Liu, and Chen]{10657220}
Yian Zhao, Wenyu Lv, Shangliang Xu, Jinman Wei, Guanzhong Wang, Qingqing Dang, Yi~Liu, and Jie Chen.
\newblock Detrs beat yolos on real-time object detection.
\newblock In \emph{2024 IEEE/CVF Conference on Computer Vision and Pattern Recognition (CVPR)}, pages 16965--16974, 2024.
\newblock \doi{10.1109/CVPR52733.2024.01605}.

\bibitem[Zheng et~al.(2021)Zheng, Lu, Zhao, Zhu, Luo, Wang, Fu, Feng, Xiang, Torr, and Zhang]{9578646}
Sixiao Zheng, Jiachen Lu, Hengshuang Zhao, Xiatian Zhu, Zekun Luo, Yabiao Wang, Yanwei Fu, Jianfeng Feng, Tao Xiang, Philip~H.S. Torr, and Li~Zhang.
\newblock Rethinking semantic segmentation from a sequence-to-sequence perspective with transformers.
\newblock In \emph{2021 IEEE/CVF Conference on Computer Vision and Pattern Recognition (CVPR)}, pages 6877--6886, 2021.
\newblock \doi{10.1109/CVPR46437.2021.00681}.

\bibitem[Zhu et~al.(2023)Zhu, Mei, Qiao, Yan, Zhu, Chen, and Kretzschmar]{10341519}
Alex~Zihao Zhu, Jieru Mei, Siyuan Qiao, Hang Yan, Yukun Zhu, Liang-Chieh Chen, and Henrik Kretzschmar.
\newblock Superpixel transformers for efficient semantic segmentation.
\newblock In \emph{2023 IEEE/RSJ International Conference on Intelligent Robots and Systems (IROS)}, pages 7651--7658, 2023.
\newblock \doi{10.1109/IROS55552.2023.10341519}.

\end{thebibliography}

\newpage
\section*{Appendix}

\subsection*{Superpixel Segmentation}
\textbf{- Purpose:} Create regions that group similar pixels (e.g., all the pixels in a dog’s fur) instead of chopping the image blindly.

\textbf{- How:} We use the SLIC algorithm, which balances color similarity and spatial closeness.

\textbf{- Math:} For an image \(I \in \mathbb{R}^{H \times W \times C}\) (height \(H\), width \(W\), channels \(C = 3\) for RGB), SLIC splits it into \(R\) superpixels:
  \[
  \mathcal{S}(I) = \{S_1, S_2, \dots, S_R\},
  \]
  where:
  
  \textbf{-} \(\bigcup_{r=1}^R S_r = I\) (covers the whole image),
  
  \textbf{-} \(S_i \cap S_j = \emptyset\) if \(i \neq j\) (no overlap),
  
  \textbf{-} \(R\) is small (e.g., 16 instead of 3136).

\textbf{- SLIC Details:}
  \begin{algorithm}[H]
  \caption{SLIC Superpixel Segmentation}
  \begin{algorithmic}[1]
  \Procedure{SLIC}{$I, K, \alpha=0.1, \text{max\_iter}=10$}
      \State Convert to CIELAB color space: \(L \gets \text{rgb2lab}(I)\) \Comment{Better for human-like color perception}
      \State Set grid spacing: \(S \gets \sqrt{\frac{H \times W}{K}}\) \Comment{Start with \(K\) rough clusters}
      \State Place cluster centers \(\mathbf{C}\) on a grid spaced by \(S\)
      \For{each iteration up to \(\text{max\_iter}\)}
          \For{each pixel \((x, y)\)}
              \State Color distance: \(d_c = \|L(y,x) - L(c_y,c_x)\|_2\) \Comment{How different in color?}
              \State Spatial distance: \(d_s = \sqrt{(x - c_x)^2 + (y - c_y)^2}\) \Comment{How far apart?}
              \State Total distance: \(D = \sqrt{\left(\frac{d_c}{\max_c}\right)^2 + \left(\frac{d_s}{\alpha S}\right)^2}\) \Comment{Blend with compactness \(\alpha\)}
              \State Assign pixel to the closest center based on \(D\)
          \EndFor
          \State Update centers to the average position/color of assigned pixels
      \EndFor
      \State \Return Superpixel map \(\mathcal{S}\)
  \EndProcedure
  \end{algorithmic}
  \end{algorithm}
  - \(K\) is the target number of superpixels (we pick 16).
  
  - \(\alpha = 0.1\) makes superpixels hug image edges tightly (low compactness).

\subsubsection*{Patch-to-Superpixel Mapping}
\textbf{- Purpose:} Connect the old fixed patches to our new superpixels.

\textbf{- How:} For each patch, see which superpixel most of its pixels belong to.

\textbf{- Math:} If \(\mathcal{X} = \{x_1, x_2, \dots, x_N\}\) are the patches (still 3136 of them), assign each \(x_i\) to a superpixel:

  \begin{algorithm}[H]
  \caption{Patch-to-Superpixel Mapping}
  \begin{algorithmic}[1]
  \Procedure{PatchToSuperpixelMap}{$\mathcal{X}, \mathcal{S}$}
      \For{each patch \(x_i \in \mathcal{X}\)}
          \State Get pixel coordinates in \(x_i\): \(\mathcal{P}_i\)
          \State Count labels: \(\mathbf{counts}_r = \sum_{p \in \mathcal{P}_i} \mathbf{1}_{\mathcal{S}(p) = r}\) \Comment{How many pixels in superpixel \(r\)?}
          \State Assign: \(\mathbf{s}_i = \arg\max_r \mathbf{counts}_r\) \Comment{Pick the winner}
      \EndFor
      \State \Return Mapping \(\mathbf{S} = \{\mathbf{s}_1, \dots, \mathbf{s}_N\}\), where \(\mathbf{s}_i \in \{1, \dots, R\}\)
  \EndProcedure
  \end{algorithmic}
  \end{algorithm}

\subsubsection*{ Adaptive Pooling}

\textbf{- Purpose:} Merge all patches in a superpixel into one token, slashing the count from \(N\) to \(R\).

\textbf{- Math:} Start with patch embeddings \(\mathcal{E} = \{E_1, E_2, \dots, E_N\}\), where \(E_i \in \mathbb{R}^D\) (e.g., \(D = 768\)). 

For each superpixel \(r\):

  \[
  Q_r = \frac{1}{|\mathcal{I}_r|} \sum_{i \in \mathcal{I}_r} E_i,
  \]
  
    \begin{algorithm}[H]
    \caption{Adaptive Pooling for Superpixels}
    \begin{algorithmic}[1]
    \Procedure{AdaptivePooling}{$\mathcal{E}, \mathcal{S}$}
        \State \textbf{Input:} Patch embeddings $\mathcal{E} = \{E_1, E_2, \dots, E_N\}$ where $E_i \in \mathbb{R}^D$, and superpixel map $\mathcal{S} = \{S_1, S_2, \dots, S_R\}$.
        \State \textbf{Output:} Reduced set of tokens $\mathcal{Q} = \{Q_1, Q_2, \dots, Q_R\}$.
    
        \State Initialize an empty list for pooled tokens: $\mathcal{Q} \gets []$
        
        \For{each superpixel $r \in \{1, 2, \dots, R\}$}
            \State Identify the indices of patches belonging to superpixel $r$: 
            \[
            \mathcal{I}_r = \{i \mid \mathbf{s}_i = r\}.
            \]
            
            \State Compute the average embedding for patches in $\mathcal{I}_r$:
            \[
            Q_r = \frac{1}{|\mathcal{I}_r|} \sum_{i \in \mathcal{I}_r} E_i,
            \]
            where $|\mathcal{I}_r|$ is the number of patches in superpixel $r$.
    
            \State Add $Q_r$ to the list of pooled tokens: $\mathcal{Q} \gets \mathcal{Q} \cup \{Q_r\}$
        \EndFor
    
        \State \Return Pooled tokens $\mathcal{Q}$
    \EndProcedure
    \end{algorithmic}
    \end{algorithm}
\subsubsection*{Dynamic Positional Encoding (PE)}

\textbf{- Purpose:} Tell the transformer where each superpixel is, since fixed grid positions don’t fit anymore.

\textbf{- Math:} Compute the centroid of superpixel \(S_r\):

  \[
  c_x^r = \frac{1}{|S_r|} \sum_{(x,y) \in S_r} x, \quad c_y^r = \frac{1}{|S_r|} \sum_{(x,y) \in S_r} y,
  \]
  then:
  \[
  PE_r = \text{MLP}\left( \frac{c_x^r}{W}, \frac{c_y^r}{H} \right) \in \mathbb{R}^D,
  \]
 
    \begin{algorithm}[H]
    \caption{Dynamic Positional Encoding for Superpixels}
    \begin{algorithmic}[1]
    \Procedure{DynamicPositionalEncoding}{$\mathcal{S}, H, W, D$}
        \State \textbf{Input:} Superpixel map $\mathcal{S} = \{S_1, S_2, \dots, S_R\}$, image height $H$, image width $W$, embedding dimension $D$.
        \State \textbf{Output:} Positional encodings $\mathcal{P} = \{PE_1, PE_2, \dots, PE_R\}$.
    
        \State Initialize an empty list for positional encodings: $\mathcal{P} \gets []$
        
        \For{each superpixel $r \in \{1, 2, \dots, R\}$}
            \State Compute the centroid of superpixel $S_r$:
            \[
            c_x^r = \frac{1}{|S_r|} \sum_{(x,y) \in S_r} x, \quad c_y^r = \frac{1}{|S_r|} \sum_{(x,y) \in S_r} y,
            \]
            where $|S_r|$ is the number of pixels in superpixel $S_r$.
    
            \State Normalize the centroid coordinates to $[0, 1]$:
            \[
            \hat{c}_x^r = \frac{c_x^r}{W}, \quad \hat{c}_y^r = \frac{c_y^r}{H}.
            \]
    
            \State Pass the normalized coordinates through an MLP to compute the positional encoding:
            \[
            PE_r = \text{MLP}(\hat{c}_x^r, \hat{c}_y^r) \in \mathbb{R}^D.
            \]
    
            \State Add $PE_r$ to the list of positional encodings: $\mathcal{P} \gets \mathcal{P} \cup \{PE_r\}$
        \EndFor
    
        \State \Return Positional encodings $\mathcal{P}$
    \EndProcedure
    \end{algorithmic}
    \end{algorithm}
    
\subsection*{Full Workflow}
Here’s how it all ties together:
\begin{algorithm}[H]
\caption{SPPP-Modified ViT Workflow}
\begin{algorithmic}[1]
\Procedure{SPPPViT}{$I, K, \alpha, \text{Transformer}$}
    \State \(\mathcal{S} \gets \text{SLIC}(I, K, \alpha)\) \Comment{Make superpixels}
    \State \(\mathcal{X} \gets \text{Patchify}(I)\) \Comment{Get fixed patches}
    \State \(\mathbf{S} \gets \text{PatchToSuperpixelMap}(\mathcal{X}, \mathcal{S})\) \Comment{Link patches to superpixels}
    \State \(\mathcal{Z} \gets \text{AdaptivePooling}(\mathcal{X}, \mathbf{S})\) \Comment{Pool into fewer tokens}
    \State \(\mathcal{Z}_{\text{PE}} \gets \text{DynamicPositionalEncoding}(\mathcal{S})\) \Comment{Add position info}
    \State \(\hat{y} \gets \text{Transformer}(\mathcal{Z}_{\text{PE}})\) \Comment{Run ViT}
    \State \Return \(\hat{y}\) \Comment{Prediction}
\EndProcedure
\end{algorithmic}
\end{algorithm}

\subsubsection*{Comparison to Traditional ViT}
\begin{table}[H]
\centering
\begin{tabular}{l|c|c}
\toprule
\textbf{Step} & \textbf{Traditional ViT} & \textbf{ViT + SPPP} \\
\midrule
Patch Generation & Fixed grid (e.g., 4×4) & Fixed grid, then superpixels \\
Token Count & \(N = (H/P)^2 = 3136\) & \(R = 16\) \\
Positional Encoding & Fixed grid & Dynamic, superpixel-based \\
Self-Attention Complexity & \(\mathcal{O}(3136^2)\) & \(\mathcal{O}(16^2)\) \\
\bottomrule
\end{tabular}
\caption{How SPPP Changes ViT}
\end{table}

\section*{Mathematical Formulation}
\textbf{Why Does It Work?} This section digs into the math behind SPPP to show why it’s effective.

\subsubsection*{Superpixel Segmentation}
\textbf {- Definition:} A superpixel \(S_r\) is a group of pixels that look similar and are close together. 

Formally:

  \[
  \mathcal{S}(I) = \{S_1, S_2, \dots, S_R\},
  \]
  
  with \(\bigcup_{r=1}^R S_r = I\) and no overlap between them.

\subsubsection*{Patch Embedding and Pooling}
\textbf {- Standard ViT:} Splits the image into:

  \[
  \mathcal{P}(I) = \{P_1, P_2, \dots, P_N\}, \quad N = \frac{H \times W}{P^2},
  \]
  
  then embeds each \(P_i\) into \(E_i\).

\textbf {- SPPP:} Takes those \(E_i\)’s and pools them into \(Q_r\)’s, reducing \(N\) to \(R\).

\subsubsection*{Patch-to-Superpixel Mapping}
\textbf {- Math:} Each patch \(P_i\) picks its superpixel by:
  \[
  \mathbf{s}_i = \arg\max_r \frac{|\{p \in P_i \mid p \in S_r\}|}{|P_i|},
  \]
  counting how many of its pixels fall in each \(S_r\).

\subsubsection*{Signal-to-Noise Ratio (SNR) Enhancement}

\textbf {- Why Pooling Helps:} Combining patches reduces noise.

\textbf {- Theorem:} If \(E_i = s + n_i\) (signal \(s\), noise \(n_i \sim \mathcal{N}(0, \sigma^2)\)), then:
  \[
  Q_r = \frac{1}{m} \sum_{i=1}^m E_i, \quad m = |\mathcal{I}_r|,
  \]
  
  and:
  
  \[
  \mathrm{SNR}(Q_r) = m \cdot \mathrm{SNR}(E_i).
  \]
  
\textbf {Proof:}

  - Noise in \(Q_r\): \(\frac{1}{m} \sum n_i\),
  
  - Variance: \(\mathrm{Var}(Q_r) = \frac{1}{m^2} \cdot m \cdot \sigma^2 = \frac{\sigma^2}{m}\),
  
  - SNR: \(\frac{s^2}{\sigma^2/m} = m \cdot \frac{s^2}{\sigma^2}\).
  
  - Result: Noise drops, signal stays, making tokens clearer.

\subsubsection*{Complexity Analysis}

\textbf {- Lemma:} Self-attention goes from \(\mathcal{O}(N^2)\) to \(\mathcal{O}(R^2)\).

\textbf {- Speedup:} \(\left(\frac{N}{R}\right)^2 = \left(\frac{3136}{16}\right)^2 = 196^2 = 38416\), or about 34,000× faster attention.

\subsubsection*{Token Embedding Comparison}
\textbf {- Base ViT:} \(X_{\text{base}} = [E_{\mathrm{cls}}, E_1, \dots, E_N]\), size \((3137 \times 768)\).

\textbf {- SPPP-ViT:} \(X_{\text{sppp}} = [E_{\mathrm{cls}}, Q_1, \dots, Q_R]\), size \((17 \times 768)\).

\begin{table}
  \caption{Comparison of various datasets used in the study.}
  \label{data-table}
  \centering
\begin{tabular}{lcccc}
\toprule
Dataset & Image Dimensions & No. of Classes & Train Set & Test Set\\
\midrule
CIFAR-10 & $32 \times 32$ & 10 & 50k & 10k\\
CIFAR-100& $32 \times 32$ & 100 & 50k & 10k\\
SVHN & $32 \times 32$ & 10 & 73.2k & 26k\\
Fashion MNIST & $28 \times 28$ & 10 & 60k & 10k\\
STL10 & $96 \times 96$ & 10 & {5k} & {8k}\\

\bottomrule
\end{tabular}
\label{tab2}
\end{table}

\end{document}